\documentclass{article}
\usepackage{spconf,amsmath,graphicx}
\usepackage{color,amssymb,multirow}
\usepackage[flushleft]{threeparttable}
\usepackage{booktabs,subfigure,kotex}
\usepackage{url}

\makeatletter
\newcommand*{\rom}[1]{\expandafter\@slowromancap\romannumeral #1@}
\makeatother
\makeatletter
\DeclareRobustCommand{\textsupsub}[2]{{%
  \m@th\ensuremath{%
    ^{\mbox{\fontsize\sf@size\z@#1}}%
    _{\mbox{\fontsize\sf@size\z@#2}}%
  }%
}}
\makeatother

\title{A Framework for Portrait Stylization with Skin-Tone Awareness and Nudity Identification}
\name{Seungkwon Kim\textsuperscript{*}, Sangyeon Kim\textsuperscript{*}, Seung-Hun Nam\textsuperscript{$\dagger$}\thanks{\textsuperscript{*} Equal contribution, \textsuperscript{$\dagger$} Corresponding author}}
\address{NAVER WEBTOON AI, Republic of Korea}

\begin{document}
\ninept
\maketitle
\begin{abstract}
Portrait stylization is a challenging task involving the transformation of an input portrait image into a specific style while preserving its inherent characteristics.
The recent introduction of Stable Diffusion (SD) has significantly improved the quality of outcomes in this field.
However, a practical stylization framework that can effectively filter harmful input content and preserve the distinct characteristics of an input, such as skin-tone, while maintaining the quality of stylization remains lacking.
These challenges have hindered the wide deployment of such a framework.
To address these issues, this study proposes a portrait stylization framework that incorporates a nudity content identification module (NCIM) and a skin-tone-aware portrait stylization module (STAPSM).
In experiments, NCIM showed good performance in enhancing explicit content filtering, and STAPSM accurately represented a diverse range of skin tones.
Our proposed framework has been successfully deployed in practice, and it has effectively satisfied critical requirements of real-world applications.

\end{abstract}

\section{Introduction}
\label{sec:intro}

The \emph{Webtoon} phenomenon has evolved beyond traditional paper-based comics.
It uses information technology to present content that is both produced and consumed in a digital format, and it is rapidly gaining global popularity.
Webtoon is thus well positioned as an optimal environment for integration with generative AI.
In this regard, portrait stylization has been an active research area, in which given individual photographs are translated into specific art styles to enhance the value of intellectual property (IP) by delivering a distinct sense of enjoyment to users \cite{back2022webtoonme}.

\begin{figure}[t]
\centering{\includegraphics[width=0.96\linewidth]{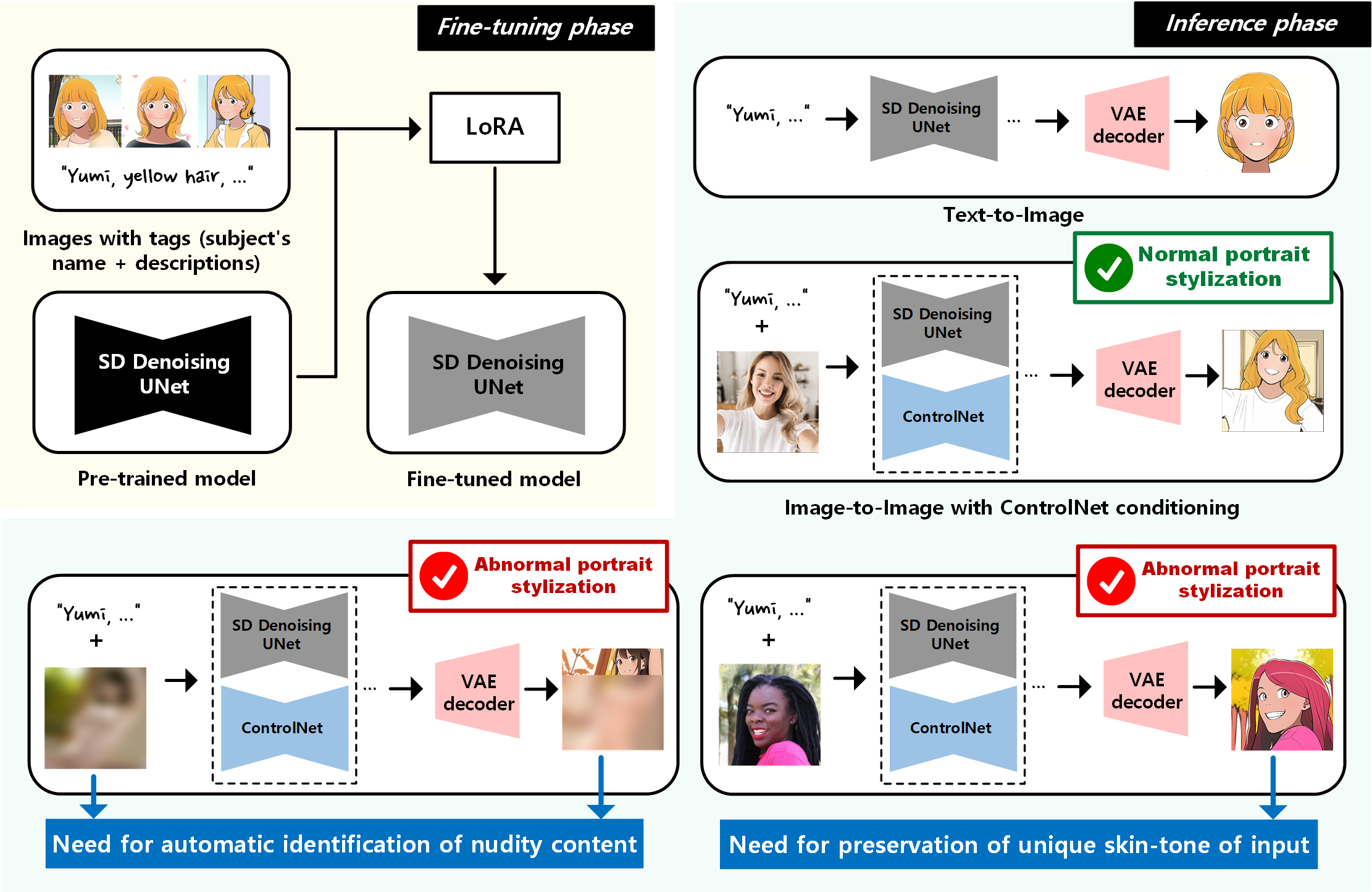}}
\vspace{-3mm}
\caption{Abnormal portrait stylization when using a generative model fine-tuned for a specific Webtoon character. The proposed framework is designed to prevent such issues.}
\vspace{-4mm}
\label{fig_overview}
\end{figure}

Various studies have investigated image-to-image (I2I)-based portrait stylization, especially those built upon StyleGAN \cite{karras2019style} and utilizing the FFHQ dataset prior \cite{pinkney2020resolution,kim2022cross,song2021agilegan}, along with techniques for expanding it to full-body portrait stylization \cite{back2022webtoonme,men2022dct}.
These approaches have provided fairly good results; however, they show some quality limitations in practical applications that make it hard to express the unique features of various IPs. Various large text-to-image (T2I) models (e.g., Stable Diffusion (SD) \cite{rombach2022high}, DALL·E 2 \cite{ramesh2022hierarchical}, Midjourney \cite{midjourney2022}) are also becoming popular nowadays. These models have been trained on large numbers of text-image pairs, and their variant models demonstrate the potential for enterprise-level applications by effectively utilizing image conditions for I2I translation \cite{rombach2022high,zhang2023adding, mou2023t2i, ye2023ip-adapter}.
Recently, studies of SD have reported quality results related to fine-tuning the whole model \cite{ruiz2023dreambooth} or only the cross-attention layer \cite{hu2021lora}, layer injection \cite{wang2020k}, and concept embedding \cite{gal2022image} for building a model to deal with a wider range of unseen styles, objects, or domains with limited data or hardware resources.
Additionally, ControlNet has been proposed to control the diffusion model based on extra conditions \cite{zhang2023adding}. Stylization using ControlNet and a fine-tuned SD model exhibits acceptable performance, as shown in the upper part of Fig.~\ref{fig_overview}.

Despite the breadth of existing studies, designing a portrait stylization framework at the business level remains challenging, as shown in the bottom part of Fig.~\ref{fig_overview}.
First, concerns exist over skin-tone expression, in which a model uniformly alters users’ actual skin tones to match those of a specific trained style, possibly leading to ethical issues.
Second, malicious users could generate sexual content with a specific style.
In IP-based businesses, safeguarding the IP is crucial; unfortunately, the neglect of this issue in existing studies could potentially lead to the reckless generation of synthesized explicit content that could severely undermine the value of IP.

\begin{figure*}[t]
\centering{\includegraphics[width=0.98\linewidth]{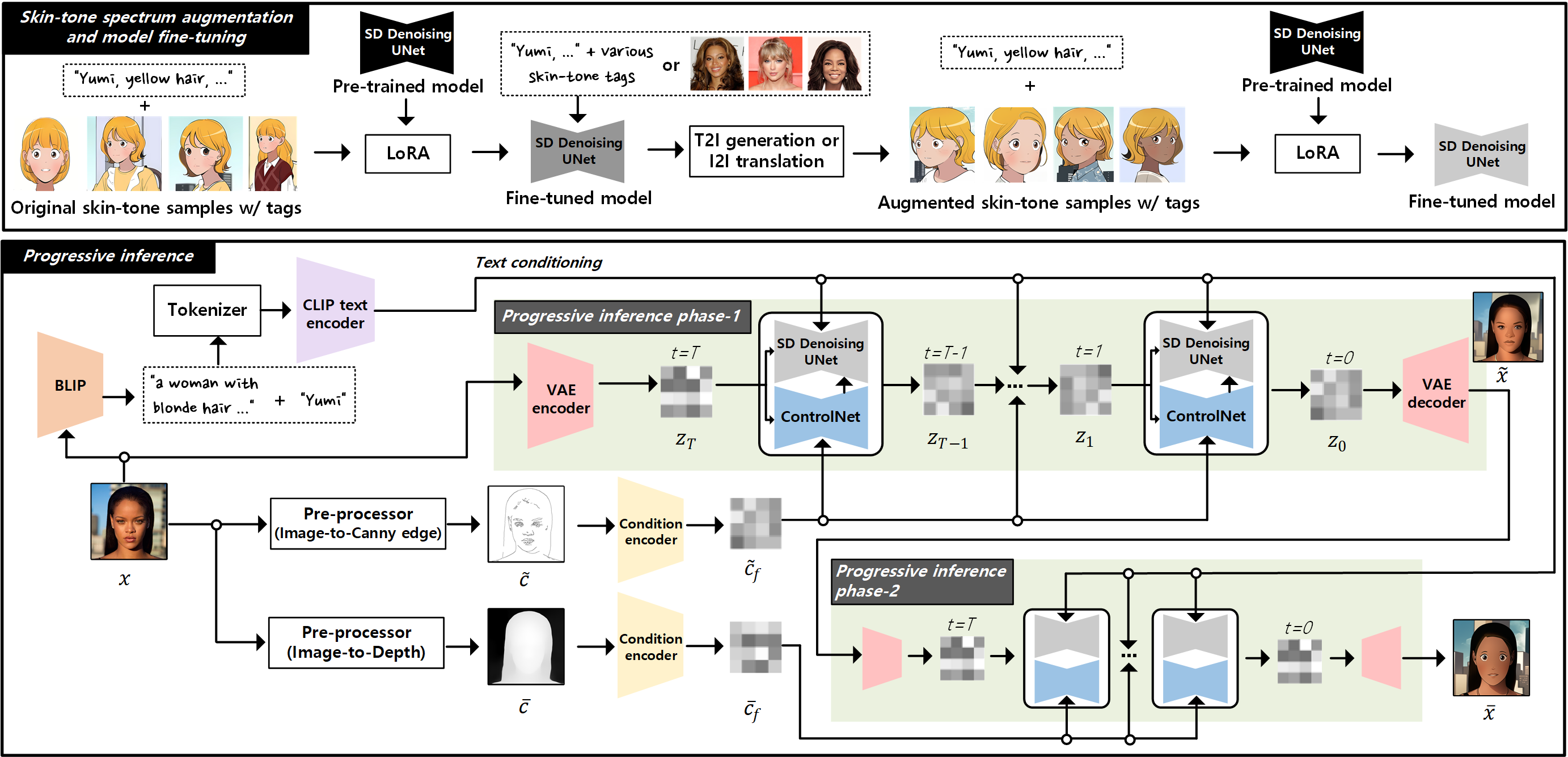}}
\vspace{-2mm}
\caption{Overall architecture of proposed STAPSM consisting of fine-tuning phase with skin-tone spectrum augmentation and progressive inference phase.}
\vspace{-4mm}
\label{fig_STAPSM}
\end{figure*}

To address these issues, we introduce a real-world application-level portrait stylization framework that can handle the diverse range of skin tones while effectively filtering explicit content. We propose the skin-tone-aware portrait stylization module (STAPSM) that contains a low-rank adaptation (LoRA) model \cite{hu2021lora} capable of not only representing a specific Webtoon character style in great detail but also mitigating skin-tone concerns (Sect.~\ref{sec:STAPSM}).
During the training phase, we use skin-tone spectrum augmentation to enhance the distribution of skin tones of the original characters, thereby refining and constructing the training dataset. Next, during the inference stage, we perform a two-stage I2I translation approach with two distinct denoising strengths and image conditions in a progressive manner.

Next, we present a nudity content identification module (NCIM) consisting of the CLIP \cite{radford2021learning} embedding classifier and BLIP \cite{li2022blip} caption-based keyword matching to identify harmful content contained in the input image (Sect.~\ref{sec:NCIM}).
We analyze the biases and lack of reliability present in existing embedding-based classifier and address them using a pre-defined keyword-based matching technique. We demonstrate that NCIM not only enhances the performance on the refined dataset, obtained from~\cite{deepghs} and categorized into normal and nudity classes through human labeling effort, but also minimizes the occurrence of inadvertent leaking of nudity images in actual service.

Our contributions are as follows.
First, our data-centric portrait stylization framework can successfully express its unique art style while preserving the input’s skin tone to a significant extent.
Second, our system ensures the generation of suitable images by involving users while also deterring misuse by malicious users through a hybrid approach of nudity filtering methods.

\section{Proposed Framework}
\label{sec:proposed_framework}
\begin{figure}[t]
\centering{\includegraphics[width=0.98\linewidth]{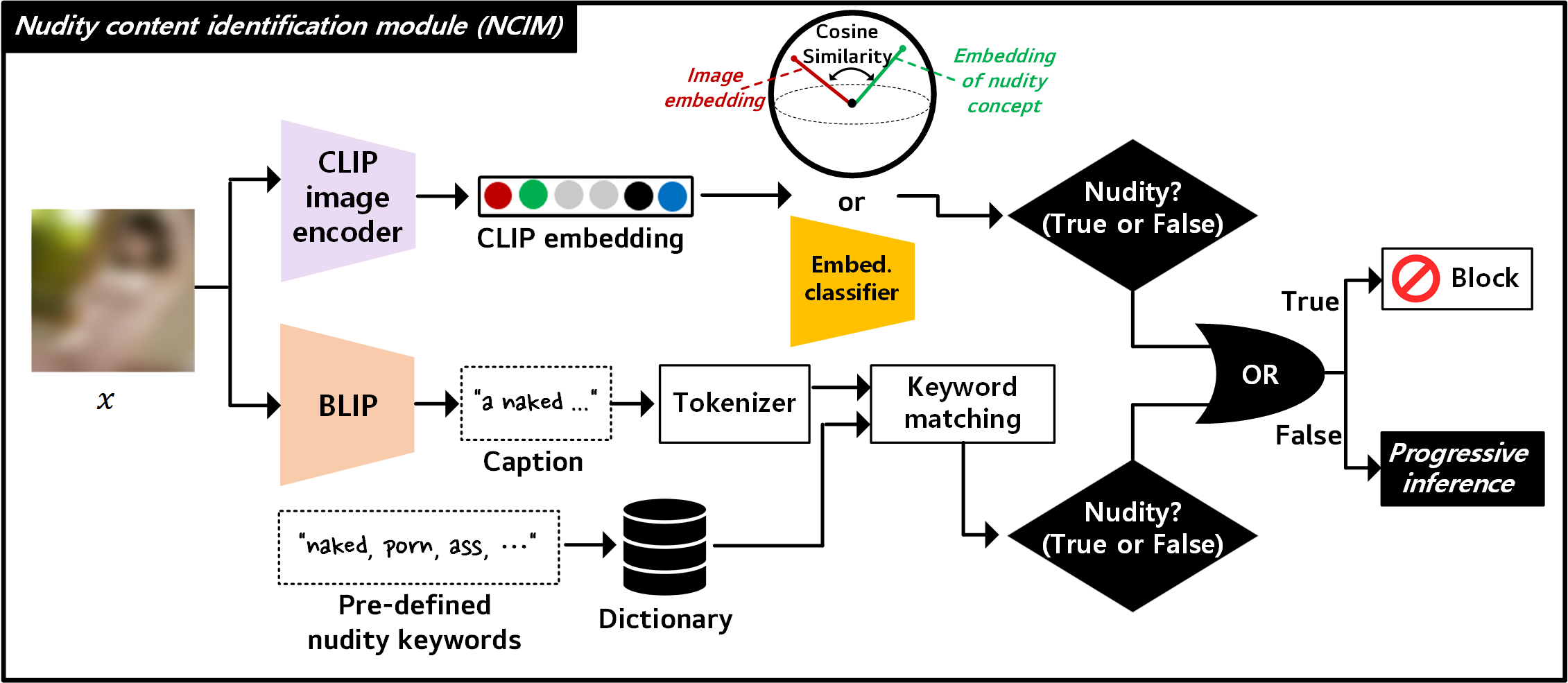}}
\vspace{-3mm}
\caption{Overall architecture of NCIM with CLIP embedding-based filtering and BLIP caption-based keyword matching.}
\vspace{-4mm}
\label{fig_NCIM}
\end{figure}

The proposed framework comprises two modules: \emph{STAPSM} for both skin-tone expression and high-quality stylization and \emph{NCIM} for identifying nudity content (see Figs~\ref{fig_STAPSM} and~\ref{fig_NCIM}).
The given portrait image in RGB format with resolution $W$$\times$$H$ is represented by $x$, where $x$ $\in \mathbb{Z}^{3\times W\times H}$.
$\tilde{c}$ and $\bar{c}$ represent the edge and depth conditions for $x$, respectively, and $\tilde{x}$ and $\bar{x}$ represent the stylization results generated through these conditions and the proposed inference approach, respectively.
$t$ and $z_{t}$ denote the time step and noisy data at $t$, respectively.

\vspace{-2mm}

\subsection{Skin-Tone-Aware Portrait Stylization Module (STAPSM)}%미팅 끝나고 이쪽부터 수정
\label{sec:STAPSM}

\begin{figure}[t]
\centering{\includegraphics[width=0.98\linewidth]{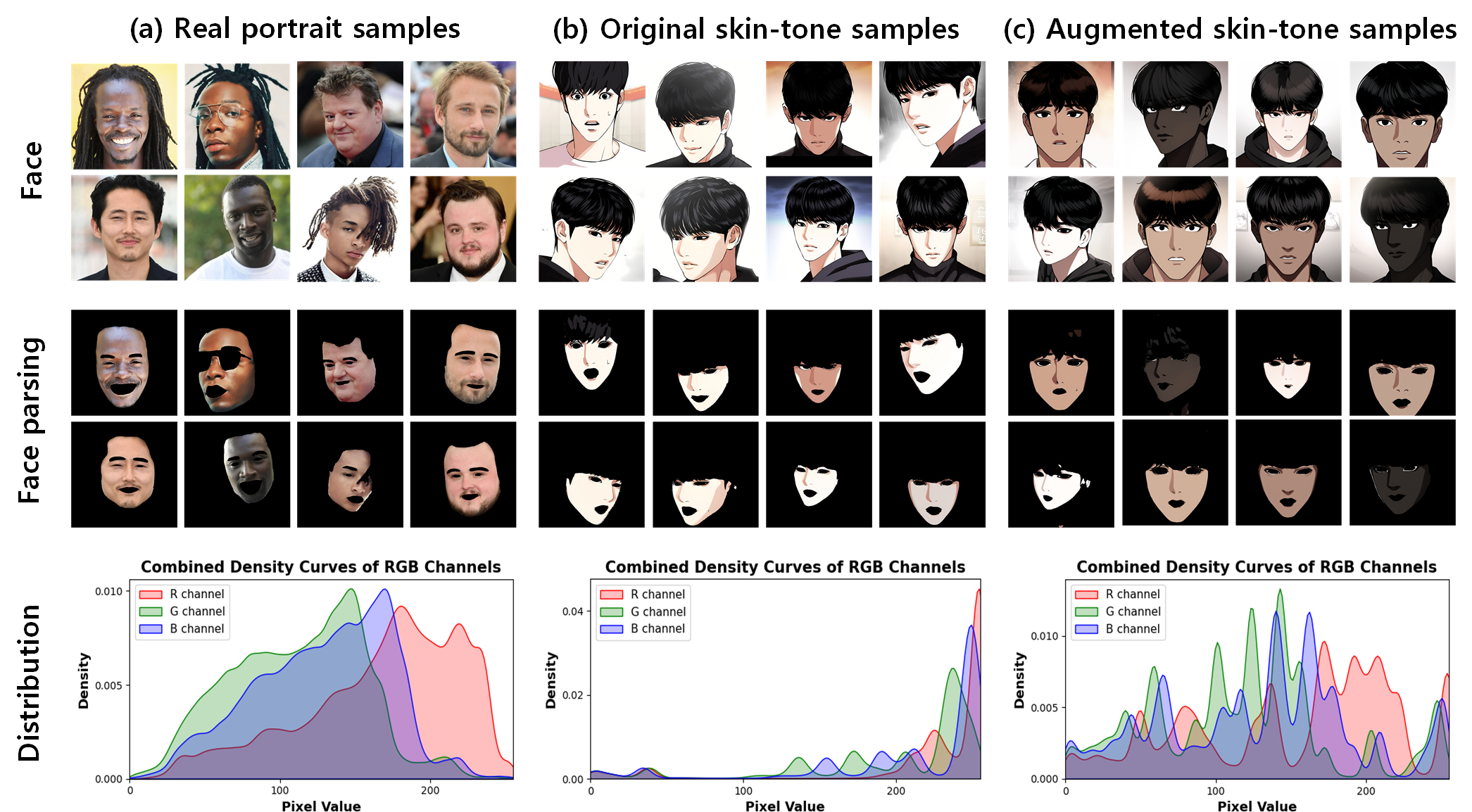}}
\vspace{-1mm}
\caption{Analysis of skin-tone distribution in real-world portrait, original, and augmented samples. Here, kernel density estimation plots for each RGB channel were generated from the parsed skin areas obtained using BiseNet v2 ~\cite{yu2021bisenet}. Instead of a distribution dominantly concentrated in specific bands as shown in (b), we adopted skin-tone spectrum augmentation to achieve the result shown in (c) to emulate an even distribution similar to the result shown in (a).}
\vspace{-3mm}
\label{fig_skintone}
\end{figure}

\begin{figure*}[t]
\centering{\includegraphics[width=0.97\linewidth]{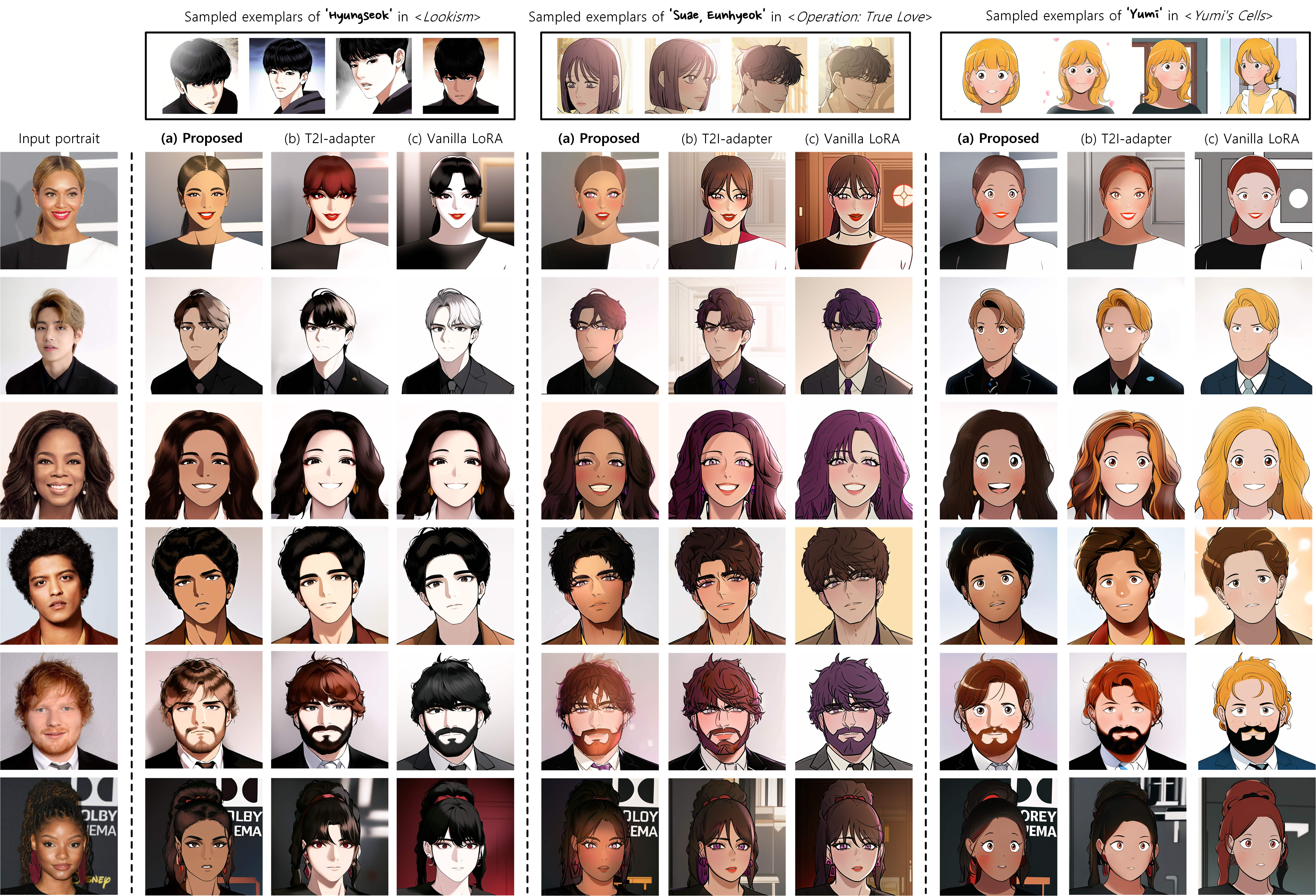}}
\vspace{-3mm}
\caption{Qualitative comparison of skin-tone representation in portrait stylization task using three Webtoon works. Compared to existing methods~\cite{mou2023t2i,hu2021lora}, our proposed method shows superior stylization quality and skin-tone expression.}
\vspace{-4mm}
\label{fig_qualitative_evaluation}
\end{figure*}

\noindent \textbf{Skin-Tone Spectrum Augmentation} The combination of ControlNet \cite{zhang2023adding} and LoRA provides high-quality results for I2I translation.
However, a limitation emerges as the LoRA model lacks explicit guidance concerning the preservation of the input skin tone.
Consequently, the generated images strictly adhere to the skin tone of a Webtoon IP character and disregard the user’s actual skin tone when used in an I2I translation scenario (see Fig.~\ref{fig_qualitative_evaluation}).
We found that this phenomenon is primarily attributable to the imbalanced distribution of character skin tones within the training dataset (see Fig.~\ref{fig_skintone}).
We also found that although LoRA generally struggles to generate skin tones beyond the character’s pre-defined range, it can produce a limited and incomplete spectrum of skin tones in response to a highly curated list of skin-tone-related prompts by using the prior color distribution of its base model.
Accordingly, we created an augmented dataset that not only preserves the unique style but also encompasses a wider range of skin colors for a character.
We achieved this by employing both T2I generation and I2I translation with two distinct input conditions: a meticulously curated set of skin-tone-related prompts\footnote{In this study, for skin-tone spectrum augmentation, we used a combination of the following skin-tone tags: \{`deepest black', `black', `dark brown', `brown', `light brown', `white', `deepest white'\} + ` skin'. To enhance the quality of augmented samples, we heuristically added parentheses and weight to the prompt (e.g., (dark brown:1.2)). For details, please contact us at shnam1520@gmail.com.} and a diverse array of portrait images (see Fig.~\ref{fig_STAPSM}).
This process ensures that the skin tone is not embedded in the dataset as a static characteristic of the character object, in a manner similar to consistent style features.
Instead, it is made a variable aspect of the style itself, and it ultimately adapts to the input skin tone during the inference stage.

\noindent \textbf{Progressive Inference} For the inference phase, we introduce a progressive two-stage inference methodology that leverages edge-based and depth-based ControlNet as shown in Fig.~\ref{fig_STAPSM}. In the initial stage, the edge-based I2I translation is executed, yielding an intermediate representation that conserves the shapes of the provided input $x$ at an edge level.
The augmented dataset significantly contributes to retaining the input skin tone (see the example of $\tilde{x}$ in Fig.~\ref{fig_STAPSM}).
However, this representation does not achieve the desired target style.
Subsequently, the depth-based I2I translation is applied to enhance the stylization quality.
In this phase, the depth map condition $\bar{c}$ is used to build upon the edge-shaped and color-preserved intermediate representation.
This approach imparts a greater degree of freedom for stylizing features based on the depth map while retaining the skin tone in the representation.
Owing to the progressive nature of the inference, it tends to preserve not only the skin tone but also intricate details pertaining to objects within the input portrait, such as clothing or letters (see Fig.~\ref{fig_qualitative_evaluation}).

\subsection{Nudity Content Identification Module (NCIM)}
\label{sec:NCIM}

Nudity content identification is a challenging task of filtering explicit or harmful content contained in the given input $x$.
Initially, nudity classification models based on supervised and unsupervised learning were dominant~\cite{nudenet}.
Recently, studies have actively focused on approaches based on CLIP~\cite{radford2021learning}, in which visual concepts are learned through natural language supervision with a variety of image and text pairs. 
For instance, SD-safety-checker (SD-SC)~\cite{sdsafetychecker} evaluates harmfulness by calculating the cosine similarity between CLIP embeddings corresponding to the pre-defined unsafe concepts and $x$, and LAION-AI-NSFW-Detector (NSFW-D)~\cite{laionnsfw} detects explicit content by using a binary classification approach in the CLIP embedding space.

\begin{figure}[t]
\centering{\includegraphics[width=0.98\linewidth]{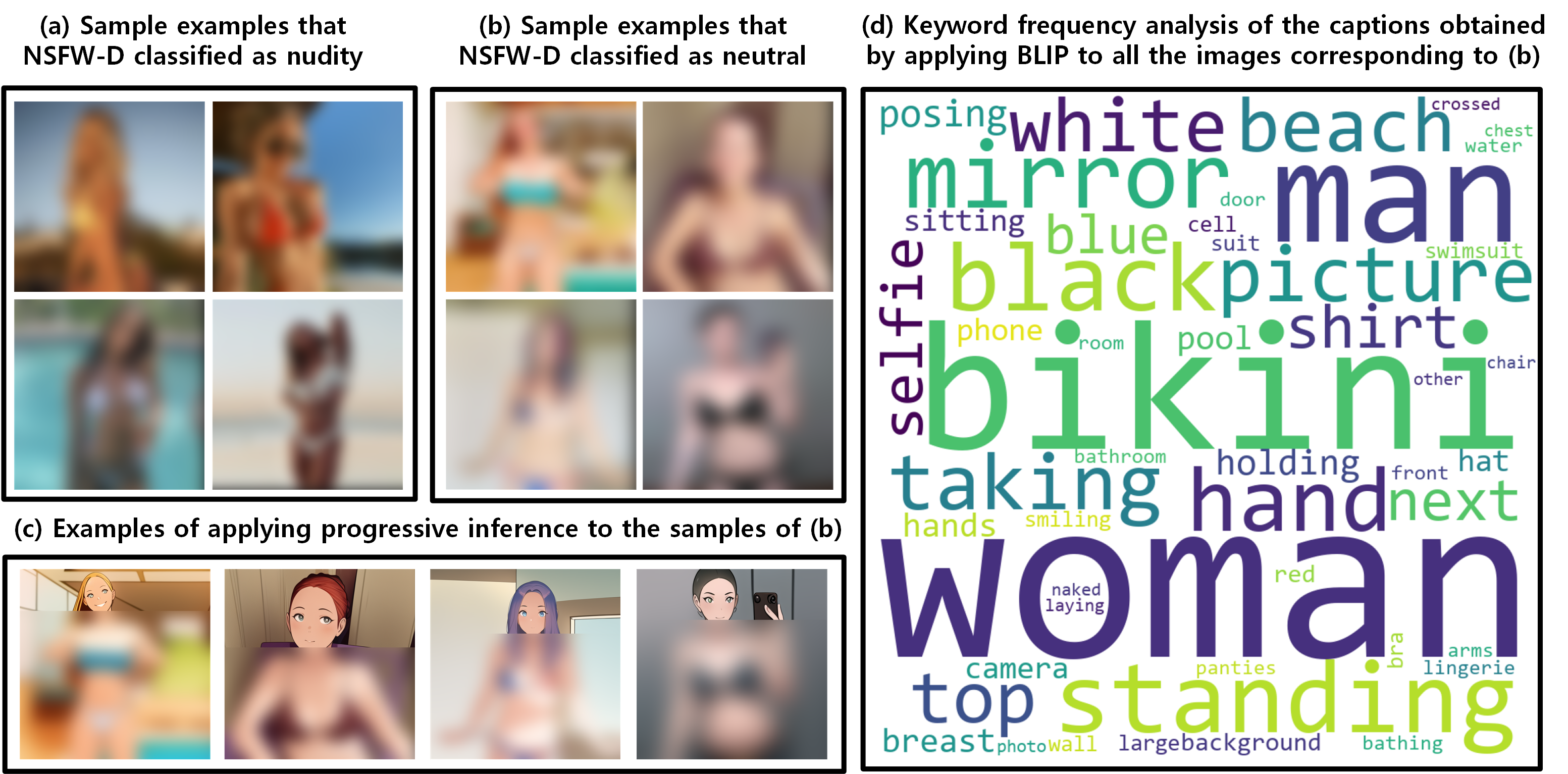}}
\vspace{-1mm}
\caption{Nudity identification analysis. NSFW-D~\cite{laionnsfw} confuses bikini images, acknowledged as explicit content in some cultures, as shown in (a) and (b). Possible misuse case when nudity filtering system does not work well, as shown in (c). Larger size of words results in a higher frequency in (d).}
\vspace{-4mm}
\label{fig_nudity_detection_analysis}
\end{figure}

However, these filters may be unreliable in real-world scenarios because they may produce varying results for the same object owing to the nature of their models.
To address this concern, we developed the NCIM in which an existing nudity filter is combined with BLIP caption-based keyword matching (BLIP-KM).
We first evaluated the BLIP image captioning results on the nudity dataset~\cite{nudenet} and found the presence of a wide range of nudity-related keywords.
Consequently, we carefully curated a pool of prompts\footnote{By using publicly disclosed harmful keywords~\cite{nsfwkeyword} and analysis of keywords for cases misjudged by existing filters (as depicted in~Fig.~\ref{fig_nudity_detection_analysis}), we constructed a dictionary consisting of more than 100 sex-related keywords (e.g., naked, nude). For details, please contact the mentioned contact email.} from among them for rule-based matching. This ensemble approach not only mitigates the biases inherent in the existing filter but also enhances filtering reliability (see Fig. \ref{fig_nudity_detection_analysis}).
By introducing the rule-based keyword matching complemented by BLIP, we can enhance nudity filtering by compensating for the edge cases missed by CLIP-model-based approaches \cite{sdsafetychecker,laionnsfw}.

\begin{table}[!t]
\caption{Quantitative evaluation of nudity content identification} % title of Table
    \centering
    \scriptsize
    \begin{tabular}{c c c c c c c}
    \hline
    \hline
    \multirow{2}{*}{\shortstack{Method}} & \multirow{2}{*}{\shortstack{Acc.}} & \multirow{2}{*}{\shortstack{Prec.}} & \multirow{2}{*}{\shortstack{Rec.}} & \multicolumn{3}{c}{F1-Score} \\
    \cmidrule(r){5-7}
    & & & & Total & Neutral& Nudity \\
    \hline
    SD-SC~\cite{sdsafetychecker} & 0.756  & 0.914 & 0.758 & 0.702 & 0.749 & 0.761 \\
    NSFW-D~\cite{laionnsfw} & 0.982  & 0.991 & 0.986 & 0.974 & 0.970 & 0.985 \\
    \hline
    BLIP-KM & 0.674  & \textbf{0.997} & 0.582 & 0.655 & \textbf{0.994} & 0.583\\
    BLIP-KM+SD-SC  & 0.844  & 0.924 & 0.871 & 0.788 & 0.747 & 0.872\\
    BLIP-KM+NSFW-D & \textbf{0.987}  & 0.990 & \textbf{0.993} & \textbf{0.980} & 0.964 & \textbf{0.992}\\
    \hline
    \hline
    \end{tabular}
    \begin{tablenotes}
    \scriptsize
    \item \textit{Notes:} Acc., Prec., and Rec.denote the abbreviations of accuracy, precision, and recall, respectively.
    \end{tablenotes}
    \vspace{-3mm}
    \label{table1}
\end{table}

\vspace{-2mm}
\section{Experiments}
\vspace{-2mm}
\subsection{Datasets and Implementation Details}

We constructed three training datasets on NAVER Webtoon works, each comprising 15 or fewer portrait images.
These were used to compare the baseline models, vanilla LoRA~\cite{hu2021lora} and T2I-adapter~\cite{mou2023t2i}. 
For our proposed method, we also augmented these datasets (Sect.~\ref{sec:STAPSM}), resulting in 40–50 portrait images for each dataset. 
As LoRA modeling was used for both the baseline and proposed models, we used the LoRA trainer~\cite{kohya} with a learning rate of $1\times10^{-4}$ and SD v1.5~\cite{rombach2022high}, trained on cartoon and character images, as the base model. During the inference stage, for our proposed model, we used two denoising strengths: $\tilde{x}=0.4$ and $\bar{x}=0.5$.
For the baseline models, we used a single denoising strength of 0.9, which we found to be effective in expressing original Webtoon styles on par with ours, along with ControlNet conditioned on a depth map.
For NCIM, we initially obtained the nudity dataset from the source~\cite{deepghs}.
Then, we refined this dataset to exclusively contain authentic images of individuals, resulting in a total of 12,252 images.
For NSFW-D~\cite{laionnsfw}, we set the threshold to 0.6 and for SD-SC~\cite{sdsafetychecker}, we used the \emph{StableDiffusionSafetyChecker} implementation of Diffusers~\cite{von-platen-etal-2022-diffusers}.

\subsection{Experimental Results}
\vspace{-2mm}
\noindent \textbf{Nudity Content Identification}
We demonstrated the effectiveness of an ensemble framework that combines existing nudity filters and the BLIP-based keyword matching method (Table~\ref{table1}).
Compared with the use of individual filters, our proposed method achieved the highest accuracy (0.987) and recall (0.993) for the explicit content category.
Our method provides the unique advantage of flexible keyword pool management, thus adapting to varying cultural norms related to sexual content in different service regions.

\noindent \textbf{Skin-tone Representation}
We compared our framework and the baseline models.
As illustrated in Fig.~\ref{fig_qualitative_evaluation}, our framework not only adeptly preserves the characters’ stylistic attributes but also faithfully reproduces the skin tone to a great extent from the input image.
In contrast, the alternative approaches provide a somewhat limited range of skin tones that cannot be suitable for practical portrait stylization services or often neglect the skin tone and express only the original character’s skin color.
We conducted a user study of 24 professionals from the Webtoon industry who answered 10 questions evaluating the similarity of the skin tone to the input image; the results indicated that our method outperforms existing approaches (Table~\ref{table2}).

\begin{table}[!t]
\caption{User study on skin-tone representation in portrait stylization} % title of Table
    \small
    \centering
    \begin{tabular}{c c c c c}
    \hline
    \hline
   Metric & Proposed & T2I-adapter \cite{mou2023t2i} & Vanilla LoRA \cite{hu2021lora} \\
    \hline
    User score& 4.31 & 3.16 & 1.53 \\
    \hline
    \hline
    \end{tabular}
    \begin{tablenotes}
    \scriptsize
    \item \textit{Notes:} Participants evaluated three methods per question, scoring 5 points for the top choice, 3 for the second, and 1 for the third, and then we averaged the score per method.
    \end{tablenotes}
    \label{table2}
    \vspace{-5mm}
\end{table}

\noindent \textbf{Real-world Application}
Our framework, branded as \emph{TOON-FILTER}, has been globally deployed as a portrait stylization service, and it has been used to produce over 2 million images for 15 popular Webtoon IPs \cite{toonfilter}.
Numerous social media users have provided positive feedback regarding its skin tone representation.
To the best of our knowledge, our framework is the first worldwide implementation of a skin-tone-aware portrait stylization service.
Additionally, the effectiveness of our nudity detection module in curbing explicit content uploads has helped protect Webtoon IPs against malicious users, and no threatening incidents have been reported in this regard.

\vspace{-2mm}

\section{Conclusion}
In this study, we proposed a portrait stylization framework that takes into account the skin-tone expression and enables more effective harmful content filtering.
Specifically, we analyzed existing skin tone biases within the training data, performed data augmentation and inference in a progressive manner, and demonstrated the ability to express skin color in a manner suitable for real-world applications.
We also analyzed the inherent biases in a widely adopted NSFW filter and subsequently developed a hybrid filtering framework that supplements the filter with an exact keyword matching method, resulting in performance improvements.
We validated the effectiveness of our proposed approach through experiments, a user study, and practical service application and demonstrated that it satisfies real-world requirements.\vfill\pagebreak

\bibliographystyle{IEEEbib}

\end{document}